\RequirePackage{amsmath}
\RequirePackage{fix-cm}
\documentclass[smallextended]{svjour3}       
\smartqed  

\usepackage{multirow}
\usepackage{makecell}
\usepackage{newtxtext,newtxmath}
\usepackage{graphicx}
\usepackage{natbib}
\usepackage{algorithm}
\usepackage[algo2e]{algorithm2e}
\usepackage{algorithmic}
\usepackage{marvosym}
\usepackage{epstopdf}

\usepackage{amssymb}

\usepackage{color}
\usepackage{colortbl}
\usepackage{hyperref}

\usepackage[inline, shortlabels]{enumitem}
\setlist{itemjoin ={,\enspace},itemjoin* = { and\enspace}}

\usepackage{booktabs}       

\usepackage{mathtools}
\usepackage{pifont}
\newcommand{\cmark}{\ding{51}}%
\newcommand{\xmark}{\ding{55}}%
\usepackage{extarrows}
\usepackage{color}
\usepackage{colortbl}
\definecolor{mygray}{gray}{.8}
\definecolor{myblue}{rgb}{0.61, 0.87, 1.0}

\usepackage{tikz}
\usepackage{etoolbox}

\newcommand{\circled}[2][]{\tikz[baseline=(char.base)]
	{\node[shape = circle, draw, inner sep = 1pt]
		(char) {\phantom{\ifblank{#1}{#2}{#1}}};%
		\node at (char.center) {\makebox[0pt][c]{#2}};}}
\robustify{\circled}

\usepackage{soul}
\soulregister\cite7 
\soulregister\citep7 
\soulregister\citet7 
\soulregister\ref7 
\soulregister\pageref7 

\newcommand*{\QEDA}{\hfill\ensuremath{\blacksquare}}

\begin{document}

\title{Probabilistic CCA with Implicit Distributions}

\titlerunning{Probabilistic CCA with implicit distributions
}

\author{Yaxin Shi \and
	    Yuangang Pan \and
	    Donna Xu \and
        Ivor W. Tsang\(^{\dagger}\) }

\institute{
              \(\dagger\) indicates the corresponding author.\\
              Yaxin Shi  \at
              Centre for Artificial Intelligence (CAI), University of Technology Sydney, Australia\\
              \email{Yaxin.Shi@student.uts.edu.au}
              \and
              Yuangang Pan  \at
              Centre for Artificial Intelligence (CAI), University of Technology Sydney, Australia\\
              \email{Yuangang.Pan@student.uts.edu.au}
              \and
              Donna Xu \at
              Centre for Artificial Intelligence (CAI), University of Technology Sydney, Australia \\
              \email{Donna.Xu@student.uts.edu.au}
              \and
              Ivor W. Tsang \at
              Centre for Artificial Intelligence (CAI), University of Technology Sydney, Australia \\
              \email{Ivor.Tsang@uts.edu.au}
}

\date{Received: date / Accepted: date}

\maketitle
\vspace{-6mm}
\begin{abstract}
Canonical Correlation Analysis (CCA) is a classic technique for multi-view data analysis. To overcome the deficiency of linear correlation in practical multi-view learning tasks, various CCA variants were proposed to capture nonlinear dependency. However, it is non-trivial to have an in-principle understanding of these variants due to their inherent restrictive assumption on the data and latent code distributions. Although some works have studied probabilistic interpretation for CCA, these models still require the explicit form of the distributions to achieve a tractable solution for the inference. In this work, we study probabilistic interpretation for CCA based on implicit distributions. We present Conditional Mutual Information (CMI) as a new criterion for CCA to consider both linear and nonlinear dependency for arbitrarily distributed data. To eliminate direct estimation for CMI, in which explicit form of the distributions is still required, we derive an objective which can provide an estimation for CMI with efficient inference methods. To facilitate Bayesian inference of multi-view analysis, we propose Adversarial CCA (ACCA), which achieves consistent encoding for multi-view data with the consistent constraint imposed on the marginalization of the implicit posteriors. Such a model would achieve superiority in the alignment of the multi-view data with implicit distributions. It is interesting to note that most of the existing CCA variants can be connected with our proposed CCA model by assigning specific form for the posterior and likelihood distributions. Extensive experiments on nonlinear correlation analysis and cross-view generation on benchmark and real-world datasets demonstrate the superiority of our model.
 
\end{abstract}

\section{Introduction}\label{sec:introduction}
Canonical Correlation Analysis (CCA)~\citep{hotelling1936relations} is a ubiquitous multi-view data analysis technique that shows promising performance in a wide range of domains, including bioinformatics~\citep{DBLP:journals/bioinformatics/GumusKSU12,naylor2010using}, computer vision~\citep{DBLP:conf/cvpr/KimWC07} and natural language processing~\citep{DBLP:conf/acl/HaghighiLBK08}. 
Assuming the data to be Gaussian distributed, the classic CCA, also known as linear CCA, simply considers linear correlation for linearly transformed data. However, in practical problems, such as multi-sensor remote sensing~\citep{ehlers1991multisensor, suri2010mutual} and medical image analysis~\citep{studholme1999overlap, bach2002kernel}, 
the structured data exhibits complex distributions. The nonlinear dependency that commonly exists in these data is insufficient to be captured with the linear CCA. 

\begin{figure}[t]
	\centering
	\hspace{-4mm}
	\includegraphics[width=11.5cm]{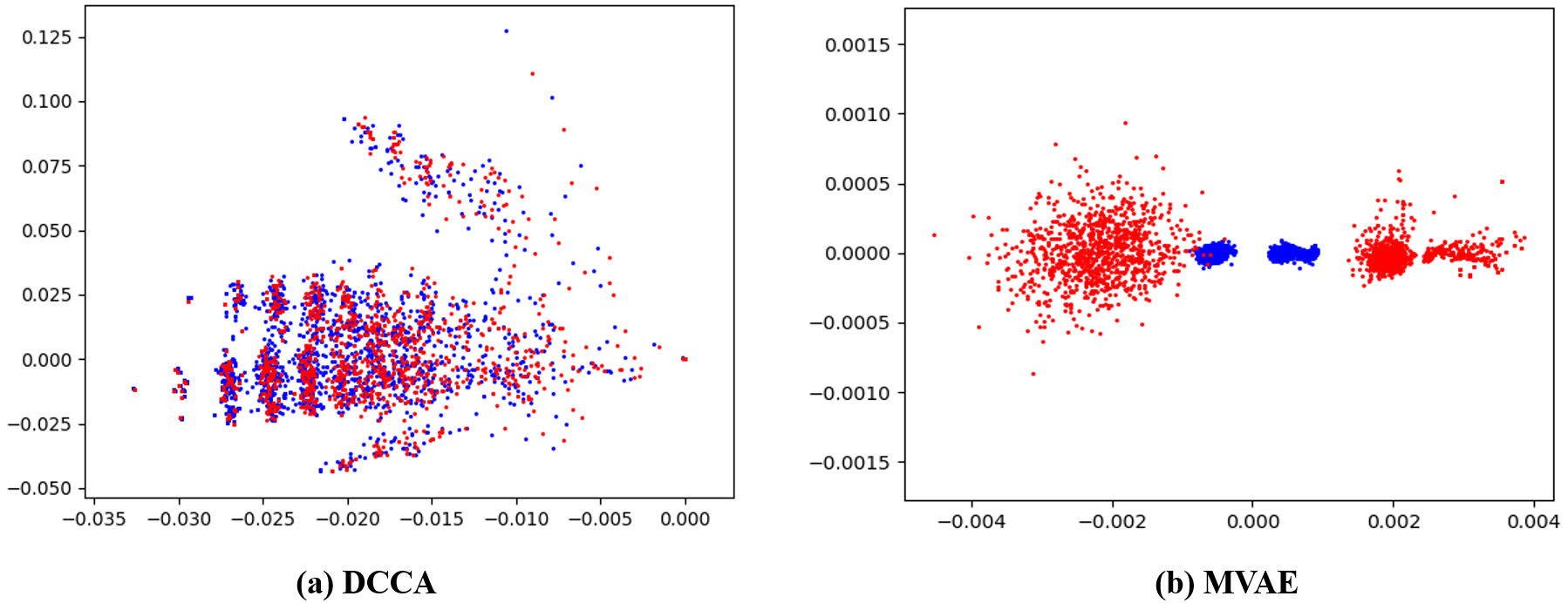}\label{fig:Case_1}
	\caption{Illustration for the latent embedding obtained with nonlinear CCA variants, DCCA~\citep{andrew2013deep} and MVAE~\citep{ngiam2011multimodal}, on the Toy dataset.} 
\end{figure}

To address this problem, a stream of nonlinear CCA variants were proposed. However, these methods still rely on restrictive assumptions on the distribution of the data and latent codes. In general, most of these nonlinear variants study nonlinear dependency from two perspectives. The first line of work studies nonlinear mapping~\citep{lai2000kernel,andrew2013deep,wang2015deep}. Nonlinear dependency is captured by preserving linear correlations for nonlinear transformed data. 
{However, the linear correlation criterion can only provide complete description for the associations when the projected data are Gaussian distributed. Considering the case shown in Fig.~\ref{fig:Case_1}.(a), the distribution of the projected data is unlikely to be Gaussian, considering the exhibited complex structure.} The other works seek to higher-order dependency metrics~\citep{vestergaard2015canonical,DBLP:journals/nn/KarasuyamaS12}.  Nonlinear dependency is achieved by capturing nonlinear dependency for linear transformed data. For these works, {although the prior for the latent space is no longer limited to be Gaussian,} the explicit form of the prior is still required to estimate the adopted nonlinear criterion~\citep{vestergaard2015canonical}. The estimation would be extremely complicated for the high dimensional data~\citep{DBLP:journals/nn/KarasuyamaS12}. There are also models that imitate nonlinear CCA by shared subspace learning via nonlinear mapping~\citep{ngiam2011multimodal}. However, the data are still assumed to be Gaussian distributed. The biased assumptions adopted in existing methods make it non-trivial to comprehend CCA from a unified perspective.

To this end, this paper {will systematically study probabilistic interpretation for CCA based on implicit distributions.} Although probabilistic interpretation for CCA has been studied in some works, they still require the explicit form of the distributions to achieve a tractable solution for the inference. Specifically, let ${x} \in \mathbb{R}^{d_{x}}$, ${y} \in \mathbb{R}^{d_{y}}$ be the random vectors in each view, and ${z} \in \mathbb{R}^{d_{z}}$ denotes the latent variable. Probabilistic~CCA (PCCA)~\citep{bach2005probabilistic} provides probabilistic interpretation for linear CCA with tractable solution, due to the \textbf{assumptions} it inherited from linear CCA: \emph{1):} the latent codes ${z}$ follows Gaussian distribution; \emph{2):} the data in each view are transformed through linear projection. Two \textbf{favorable conditions} are satisfied with these assumptions: \emph{1).}~$p({x},{y}|{z})$ can be modeled with the joint covariance matrix, with which the conditional independent constraint for CCA~\citep{drton2008lectures} can be easily imposed.
\begin{equation}
	p{({x},{y}|{z})} = p{({x}|{z})}p{({y}|{z})}.\label{eq:conditional_independent}
\end{equation}
\emph{2).} Due to the conjugacy of the prior and the likelihood, the posterior, i.e. \begin{small}$p{({z}|{x},{y})} = \frac{p{({x},{y}|{z})}p({{z}})}{p{({x},{y})}}$\end{small} can be presented with an analytic solution~\citep{tipping1999probabilistic}.
These two conditions facilitate a valid probabilistic interpretation for linear CCA. Probabilistic interpretation for nonlinear CCA models is much more difficult to achieve, since both the two conditions are violated. Specifically, as nonlinear dependency is to be captured, the Gaussian assumption made on the prior is violated. Therefore, the linear correlation is no longer an ideal criterion for the problem. Furthermore, as the mappings to the latent space are complex in some nonlinear variants, conjugacy between the prior and the likelihood is violated. The intractable posterior distributions make it hard to enforce the conditional independent constraint (Eq.~\eqref{eq:conditional_independent}) for the problem. Although deep variational CCA (VCCA)~\citep{wang2016deep} is proposed to interpret nonlinear CCA, it still adopts a Gaussian distribution assumption on the posterior to achieve a tractable solution for the inference of multi-view data with KL-divergence. Besides, as there is no explicit connection between its objective and that of CCA, the relevancy of VCCA and CCA is still unclear.

Considering the aforementioned challenges, in this paper, we provide probabilistic interpretation for CCA with Conditional Mutual Information (CMI). We present minimum CMI as a new criterion for CCA to consider both linear and nonlinear dependency. To estimate CMI without the explicit form of the distributions, we derive an objective which can provide an estimation for CMI with inference methods. To facilitate efficient Bayesian inference for multi-view analysis with implicit distributions, we propose Adversarial CCA (ACCA), which achieves consistent encoding for multi-view data with the consistent constraint imposed on the marginalization of the implicit posteriors. Most of the existing CCA variants can be connected with our model based on certain assumptions on the posterior and likelihood distributions. We demonstrate the superior consistency achieved with our model contributes to better nonlinear correlation analysis and cross-view generation performance on both benchmark and real-world datasets. The contributions of this work can be summarized as follows:

\begin{itemize}
	
	\item [1.] We present minimum Conditional Mutual Information (CMI) as a new criterion for CCA, which can consider both linear and nonlinear dependency for CCA without additional concern for the conditional independent constraint.
	\vspace{1mm}
	\item [2.] We provide an objective which can provide an estimation for CMI in multi-view learning, without the explicit form of the distributions.
	\vspace{1mm}
	\item [3.] We propose Adversarial CCA (ACCA) which facilitates Bayesian inference for multi-view analysis with implicit distributions. Most of the existing CCA variants can be explained with our model with specific distribution assumptions. 
	\vspace{1mm}
	\item [4.] Achieving superior consistency for the encoding of multi-view data, our proposed ACCA presents superior performance in nonlinear dependency analysis and cross-view data generation task.
	
\end{itemize}

The remainder of this paper is organized as follows. Section~\ref{sec:related_work} discusses the related works and preliminaries of our study. In Section~\ref{sec:ICCA}, we present Conditional Mutual Information as a new criterion for CCA and provide the objective for efficient estimation for CMI. Section~\ref{sec:ACCA} presents our design of Adversarial CCA (ACCA) and explains its connection with existing CCA variants. Section~\ref{sec:experiments} demonstrates the superiority of our model through empirical results on both synthetic and real-world datasets. Section~\ref{sec:conclusions} concludes the paper and envisions the future work. 

\begin{table}[t]
	\begin{center}
		\renewcommand{\arraystretch}{1.2}
		\caption{
			Comparison of existing CCA variants. Linear methods are marked with grey, while others are nonlinear extensions; our method is marked with blue. The column 5 indicates whether the method avoids Gaussian distribution assumption on $p({z})$. The column 6-8 indicates whether the generative models can handle corresponding distribution with implicit form.}
		\label{tab:CCAvariants}
		\setlength{\tabcolsep}{1.1mm}{
			\scalebox{0.94}{
				\begin{tabular}{c|cc|c|c|ccc}
					\toprule				
					\multirow{2}{*}{\textbf{Methods}} & \multicolumn{2}{c|}{\textbf{Nonlinear}} & \multirow{2}{*}{\textbf{Generative}} & \multirow{2}{*}{\begin{tabular}[c]{@{}c@{}}\textbf{Avoids Gaussian}\\  \textbf{distribution on $p(\mathbf{z})$}\end{tabular}} & \multicolumn{3}{c}{\textbf{Implicit}} \\ \cline{2-3}
					\multicolumn{1}{c|}{} & \multicolumn{1}{c}{\textbf{Mapping}} & \multicolumn{1}{c|}{\textbf{Criteria}} & & & $p(\mathbf{z})$ & $p(\mathbf{x},\mathbf{y}|\mathbf{z})$ & $p(\mathbf{z}|\mathbf{x},\mathbf{y})$ \\ \hline
					\rowcolor{mygray}CCA      & \xmark    & \xmark   & \xmark & \xmark & \xmark & \xmark & \xmark\\
					\rowcolor{mygray}PCCA     & \xmark    & \xmark   & \cmark & \xmark & \xmark & \xmark & \xmark\\
					KCCA     & \cmark     & \xmark      & \xmark & \xmark & \xmark & \xmark & \xmark \\
					DCCA     & \cmark     & \xmark      & \xmark & \xmark & \xmark & \xmark & \xmark \\
					CIA      & \xmark     & \cmark      & \xmark & \cmark & \xmark & \xmark & \xmark \\
					LSCDA    & \xmark     & \cmark      & \xmark & \cmark & \xmark & \xmark & \xmark\\
					MVAE    & \cmark      & \textbf{-}      & \xmark & \xmark & \xmark & \xmark & \xmark\\
					VCCA     & \cmark     & \textbf{-}  & \cmark & \xmark & \xmark & \cmark & \xmark\\
					Bi-VCCA  & \cmark     & \textbf{-}  & \cmark & \xmark & \xmark & \cmark & \xmark\\
					\rowcolor{myblue}ACCA & \cmark      & \cmark & \cmark & \cmark & \cmark & \cmark & \cmark\\
					\bottomrule
		\end{tabular}}}
	\end{center}
	\vspace{-4mm}
\end{table}

\section{Related work and preliminaries}\label{sec:related_work}
In this section, we give a review of the existing works that are related to our study.

\vspace{0.5mm}
\textbf{Canonical Correlation Analysis (CCA)}: CCA~\citep{hotelling1936relations} is a powerful statistic tool for multi-view data analysis. Let $X = \{x^{(i)}, y^{(i)}\}_{i=1}^{N}$ consists $N$ i.i.d. samples with pairwise correspondence in multi-view scenario, the classic CCA aims to find linear projections for the two views, ($W_{x}^{'}X, W_{y}^{'}Y$), such that the linear correlation between the projections are mutually maximized, namely
\begin{equation}
	\max {\quad} \rho = corr\{W_{x}^{'}{X}, {W}_{y}^{'}{Y}\}=\frac{{W}_{x}^{'}\mathbf{\Sigma}_{xy}{W}_{y}}{\sqrt{{W}_{x}^{'}\mathbf{\Sigma}_{xx}{W}_{x}{{W}_{y}^{'}\mathbf{\Sigma}_{yy}{W}_{y}}}},\label{eq:linear_CCA}
\end{equation} 
where $\rho$ denotes the correlation coefficient, ${\mathbf{\Sigma}_{xx}}$ and  ${\mathbf{\Sigma}}_{yy}$ are the covariance of $X$ and $Y$, respectively, ${\mathbf{\Sigma}}_{xy}$ denotes the cross-covariance of $X$ and $Y$. 

Assuming the data to be Gaussian distributed, the classic CCA simply captures linear correlation under linear projections. It is often insufficient to analyse complex real-world data that exhibit higher-order dependency~\citep{suzuki2010sufficient}.

\vspace{0.5mm}
\textbf{Nonlinear CCA variants}: Various nonlinear CCA variants were proposed to capture nonlinear dependency in multi-view problems. 
Most of these models can be grouped into two categories according to the extension strategy (see Table~\ref{tab:CCAvariants}). For the first category, the nonlinear extension is conducted by capturing linear correlation for nonlinear transformed data, e.g. Kernel CCA~(KCCA)~\citep{lai2000kernel} and Deep CCA~(DCCA)~\citep{andrew2013deep}. The adopted linear correlation is optimal only when the common latent space is Gaussian distributed. However, since nonlinear mappings are adopted, the posterior is intractable, in which the Gaussian distribution assumption can not be fulfilled. For the other category, the extension is conducted by capturing high-order dependency for linear transformed data. Most of the works adopt mutual information or its variants as the nonlinear dependency measurement, e.g. Canonical Information Analysis~(CIA)~\citep{vestergaard2015canonical} and Least-Squares Canonical Dependency Analysis (LSCDA)~\citep{DBLP:journals/nn/KarasuyamaS12}, respectively. However, the explicit form of the latent distribution is required to estimate the adopted criterion in these methods. The estimation would be extremely complicated for the high dimensional data~\citep{kraskov2004estimating}. There are also works that adopt Multi-View AutoEncoders~(MVAE)~\citep{ngiam2011multimodal} to discover the correlation of the multi-view data via multi-view reconstruction. However, it still assumes the latent space to be Gaussian distributed, which is often violated practically (see Fig.~\ref{fig:Case_1}.(b)).     Furthermore, since there is no explicit objective for the model as discovering correlations, its connection with CCA is unclear.
\begin{figure}[t]
	\centering
	\includegraphics[width=11cm]{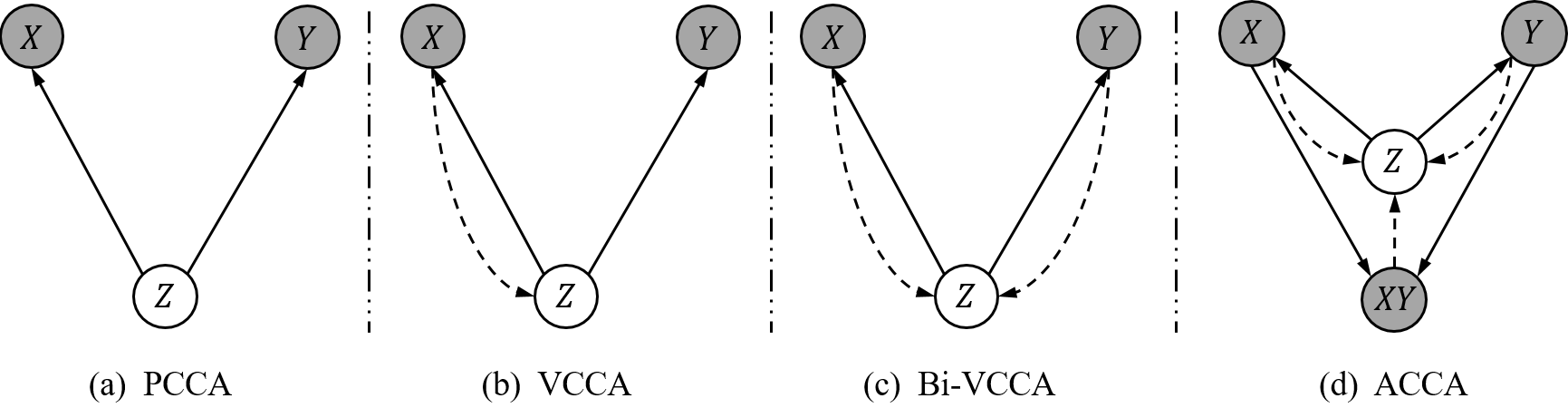}
	\caption{Graphical diagrams for generative nonlinear CCA variants. {The solid lines in each diagram denote the generative models $p_{\theta}({z})p_{\theta}({*}|{z})$. The dashed lines denote the variational approximation $q_{\phi}({z}|{*})$ to the intractable posterior $p_{{\theta}}({z}|{*})$.} }\label{fig:Graphics} \vspace{-4mm}
\end{figure}

\textbf{Probabilistic interpretation for CCA:} To deepen the understanding of CCA-based models, some works attempt to study a probabilistic interpretation for CCA. 

PCCA~\citep{bach2005probabilistic} studies the probabilistic interpretation for linear CCA (Eq.~\eqref{eq:linear_CCA}). It states that maximum likelihood estimation for the model in Fig.~\ref{fig:Graphics}.(a), leads to the canonical correlation directions, by
\begin{gather}\label{eq:PCCA}
	{z}\sim\mathcal{N}({0},{I}_{d}),\nonumber\\
	{x}|{z}\sim \mathcal{N}({W}_{x}{z}+ {\mu}_{x},{\Phi_{x}}),\\
	{y}|{z}\sim\mathcal{N}({W}_{y}{z}+{\mu}_{y},{\Phi_{y}}),\nonumber
\end{gather}
where $d$ denotes the dimension of the projected space. Obviously, inheriting from the linear CCA, PCCA adopts two assumptions: \emph{1).} the latent space follows Gaussian distribution, with which the linear correlation is a ideal criterion; \emph{2).} the data in each view are transformed through linear projection. As the prior and the likelihood are conjugate, $p({x},{y}|{z})$ can be modeled with the joint covariance matrix, with which the conditional independent constraint~\citep{drton2008lectures} can be easily imposed.

However, there lacks a probabilistic interpretation for nonlinear CCA variants that can explain the nonlinear dependency captured by these models. An intuitive idea is to generalize PCCA to these nonlinear models. However, PCCA cannot interpret nonlinear CCA variants for mainly two reasons: \emph{1).} The correlation criterion adopted in PCCA cannot capture the high-order dependency between the variables. \emph{2).} The conditional independent constraint is hard to enforce in the optimization of nonlinear CCA variants. In~\citep{wang2016deep}, deep variational CCA (VCCA) is presented, which is claimed to provide interpretation for the first category of nonlinear CCA models parametrized by DNNs. In addition, a variant named bi-VCCA is further presented, which incorporates encoding for both the two views. However, the connection between their objectives and nonlinear dependency is not clear. Furthermore, the prior $p({z})$ and the approximate posteriors $q({z}|{x},{y})$ are still required to be Gaussian distributions so that the KL-divergence can be computed analytically. Since data are usually complex in real-world tasks, simple Gaussian prior restricts the expressive power of these CCA models~\citep{mescheder2017adversarial}.

Consequently, there are two challenges to consider probabilistic interpretation for nonlinear CCA models. First, a new criterion which can measure nonlinear dependency with implicit distributions is required. Second, the conditional independent constraint needs to be easily imposed with the proposed criterion. 
\vspace{-3mm}
\section{Probabilistic interpretation for CCA with CMI}\label{sec:ICCA}

In this section, we present a probabilistic interpretation for CCA with Conditional Mutual Information~(CMI). First, we present the minimum CMI as a new criterion for CCA in Sect.~\ref{sec:section_CMI}. Then, we provide our objective for the estimation of CMI in Sect.~\ref{sec:section_ICCA}. 

\vspace{-3mm}
\subsection{Minimum CMI: a new criterion for CCA }\label{sec:section_CMI}

Given three random variables ${X},{Y}$ and ${Z}$, the CMI measures conditional dependency, i.e. mutual information between ${X}$ and ${Y}$ given the variable ${Z}$~\citep{zhang2014conditional}. 
\begin{equation}\label{eq:CMI}
	I{({X};{Y}|{Z})}=\iiint 
	p{({z})} p{({x},{y}| {z})} \log \frac{p{({x},{y}|{z})}}{p{({x}|{z})}p{({y}|{z})}} d{z}d{x}d{y} 
\end{equation} 
In general, CMI owns two properties: 
\begin{itemize}
	\item[\circled{\textbf{1}}] It is a general measurement that can evaluate both linear and nonlinear dependency between the variables~\citep{bertran2019learning}.
	\vspace{1mm}
	\item[\circled{\textbf{2}}]  $I({X};{Y}|{Z})\geq 0$~\citep{cover2012elements}.
\end{itemize}

\begin{figure}
	\begin{center}
		\hspace{3mm}
		\includegraphics[width=10cm]{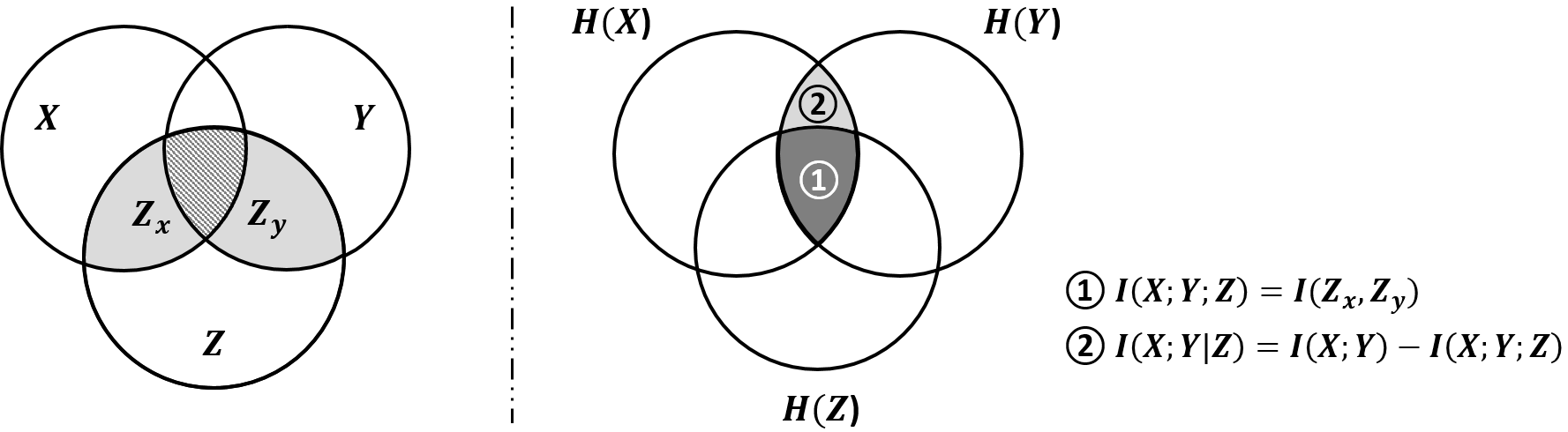}
	\end{center}
	\caption{The left side presents the basic diagram of CCA based on the variable. The right side shows the Venn diagram of the information theoretic measurements for ${X}$, ${Y}$ and ${Z}$ in CCA.}\label{fig:CMI}
\end{figure}

The connection between CMI and the general objective of CCA can be illustrated with Fig.~\ref{fig:CMI}. Specifically, from the probabilistic perspective, given the multi-view data (${X},{Y}$), CCA aims to find latent variable ${Z}$ with which the dependency between ${X}$ and ${Y}$ can be maximumly captured. Adopting mutual information as the dependency measurement, the mutual information between the latent codes ${Z}_{x}$ and ${Z}_{y}$ is to be maximized. Since $I{({X};{Y})}$ is constant for a given problem, $I{({X};{Y}|{Z})}$ is to be minimized for CCA. This makes the minimum CMI a new criterion for CCA models. 

The proposed minimum CMI criterion overcomes the challenges concluded in Sect.~\ref{sec:related_work} simultaneously. First, CMI can measure the nonlinear dependency~\citep{rahimzamani2017potential} between two random variables, making it a competent statistic to interpret nonlinear CCA variants. Second, the conditional independent constraint for CCA-based models is automatically satisfied with the minimum CMI, since the optimal, $I({X};{Y}|{Z}) = 0$~\citep{cover2012elements}, is achieved with Eq.~\eqref{eq:conditional_independent}.
 \vspace{-3mm}\\
\textbf{Discussion}: As shown with Fig.\ref{fig:CMI}, in the  multi-view learning scenario, CMI can be interpreted as the reduction of mutual information between the two views when mapped into the common latent space. The superiority of minimum CMI as CCA criterion can be explained from the following two aspects: 
\begin{itemize}
	\item [1).] Compared with the linear correlation~\citep{taylor1990interpretation}, CMI is a more general criterion that can interpret existing CCA models without simplifying distribution assumptions.
	\vspace{1mm}
	\item [2).] Compared with $I({Z}_{x},{Z}_{y})$ adopted in ~\citep{vestergaard2015canonical} and~\citep{DBLP:journals/nn/KarasuyamaS12}, CMI presents explicit connection with the conditional independent assumption in CCA, which makes it a better criterion to consider nonlinear dependency for CCA. 
\end{itemize}
\begin{remark}As a variant of mutual information, CMI considers information about all the dependency~\citep{gelʹfand1959calculation}, both linear and nonlinear, in CCA without explicit assumptions on the mapping and the distribution of the latent space. Existing CCA models can all be explained with this criterion as follows. 
	
The classic linear CCA can be explained with CMI since there exists an exact relationship between linear correlation and CMI (refer to Eq.\eqref{eq:CMI_rho})when both the original data and the latent codes are assumed to be Gaussian distributed~\citep{brillinger2004some}. 
		\begin{equation}
		I({X};{Y}|{Z}) = I{({X},{Y})} - I{({Z}_{x},{Z}_{y})} = -\frac{1}{2} \log \frac{(1-\rho_{XY}^{2})}{(1-\rho_{Z_xZ_y}^{2})}, \label{eq:CMI_rho}
		\end{equation}
		where $\rho_{XY} = corr\{{X},{Y}\}$ and $\rho_{Z_x Z_y} = corr\{{Z}_x, {Z}_y\}$.
		
		The nonlinear CCA variants (listed in Table~\ref{tab:CCAvariants}) can be explained with CMI, since CMI eliminates the restriction on the form of mapping and the distribution of the common latent space.
 \QEDA
\end{remark}

\subsection{Objective of the estimation for CMI}\label{sec:section_ICCA}

To provide an estimation for CMI without the explicit form of the distribution, we derive an objective which can provide an estimation for CMI with efficient inference methods.

Based on the definition of CMI, i.e. Eq.~\eqref{eq:CMI}, we have 
\begin{eqnarray}
	&&I_{\theta}{({X};{Y}|{Z})}\nonumber \\ 
	&=&\iiint 
	p{({z})} p{({x},{y}|{z})} \log \frac{p{({x},{y}|{z})}}{p{({x}|{z})}p{({y}|{z})}} d{z}d{x}d{y} \nonumber \\
	&=& H(X, Y) + \mathbb{E}_{p({x},{y})} [D_{KL}(p_{\theta}({z}|{x},{y})\parallel p({z}))- \mathbb{E}_{p_{\theta}({z}|{x},{y})} [\log {p_{\theta}({x}|{z})}+\log {p_{\theta}({y}|{z})}]  \nonumber 
\end{eqnarray}
The $H(X,Y)$ is a constant and has no effect on the optimization. Consequently, the minimum CMI criterion can be achieved by minimizing the remaining terms, namely 

\begin{equation}\label{eq:ICCA_FN}
	\min\limits_{\theta} \mathbb{E}_{p(x,y)}~[F(\theta;x,y)]  \simeq \frac{1}{N} \sum_{i = 1}^{N} F(\theta;x^{(i)},y^{(i)}).
\end{equation}
where $
F(\theta;x,y)=-\mathbb{E}_{p_\theta({z}|{x},{y})}~[\log {p_\theta({x}|{z})}+\log {p_\theta({y}|{z})}] +D_{KL}(p_\theta({z}|{x},{y})\parallel p({z}))\label{eq:ICCA_CMI_Fp}.$

Although Eq.~\eqref{eq:ICCA_FN} avoids the difficulties
in direct estimation CMI, the objective is still hard
to optimize since the posterior $p_\theta(z|x,y)$ is unknown or intractable for practical multi-view learning problems. 
Consequently, efficient Bayes inference technique is to be adopted for an estimation for CMI based on implicit distributions. 

\section{Adversarial CCA}\label{sec:ACCA}
In this section, we present our proposed Adversarial CCA (ACCA), which facilitates efficient Bayesian inference for multi-view analysis based on implicit distributions. We first state the deficiency of existing generative CCA variants in cross-view analysis in Sect.~\ref{sec:ACCA_deficiency}. Then, we give our motivation for ACCA in Sect.~\ref{sec:ACCA_motivation}. Next, we present the technical details, including the model design, the formulation and training details of ACCA in Sect.~\ref{sec:ACCA_design}, Sect.~\ref{sec:ACCA_formulation} and Sect.~\ref{sec:ACCA_training} respectively. In the end, we discuss the connection between  ACCA and other probabilistic CCA variants 
in Sect.~\ref{sec:ACCA_connections}.

\vspace{-5mm}
\subsection{Deficiency of CCA variants in cross-view analysis}~\label{sec:ACCA_deficiency}
As a powerful technique for multi-view data analysis, CCA and its variants have been widely adopted in discriminant learning tasks~\citep{kim2007discriminative,li2018survey}. However, seldom attention has been paid on the generative CCA model which is capable to handle cross-view data analysis task. Considering we have images of different angles or views for an object, it would be promising to generate one view of the data given single input from the other view, namely cross-view data generation~\citep{regmi2018cross}. The application would be even more interesting for multimodal data, e.g. pairwise visual and audio data. The study can also benefit the common partial view problem~\citep{li2014partial,shi2019label} in multi-view data analysis tasks. 

Alignment of the multi-view data plays a crucial role in the aforementioned cross-view generation task~\citep{ngiam2011multimodal}. There are two challenges to achieve superior alignment in these tasks. First, handling implicit distributions would greatly benefit the expressiveness for the encodings of each view~\citep{mescheder2017adversarial}, but how to achieve a tractable solution for implicit posterior remains a problem. Second, how to achieve consistency for the implicit posteriors obtained with each view.

Existing generative CCA variants suffer from misalignment in these tasks due to their biased assumption on the posterior distribution. Specifically, since PCCA adopts linear mapping on Gaussian distributed data, it can directly achieve an analytic solution for the posterior (discussed in Section.~\ref{sec:introduction}). Bi-VCCA also assumes the posterior to be Gaussian distributed to achieve tractable solutions for the inference of each view. However, it adopts a heuristic combination of KL-divergence to achieve consistency of the posteriors, which yields an improper inference for multi-view analysis. Consequently, Bi-VCCA still suffers from the misaligned encoding for the two views.
\vspace{-8mm}
\subsection{Motivation} \label{sec:ACCA_motivation}
To tackle the aforementioned challenges in the cross-view analysis tasks, we derive a tractable solution for implicit posterior based on the first principle of Bayesian inference---marginalization. Since the marginalization on $p_\theta({z}|{x},{y})$ in Eq.\eqref{eq:ICCA_CMI_Fp}, i.e. $\iint p_{\theta}({z}|{x},{y})\,p({x},{y})d{x}d{y}$, is intractable, {we introduce an approximation for the exact posterior with a tractable posterior $q(z|x,y)$ such that a tractable solution can be achieved for Eq.~\eqref{eq:ICCA_FN} with
\begin{align}\label{eq:ICCA_F_q}
	&\lefteqn{\min \frac{1}{N} \sum_{i = 1}^{N} F(\theta,\phi; {x}^{(i)},{y}^{(i)})}\\ 
	& \; = -\frac{1}{N} \sum_{i = 1}^{N} \mathbb{E}_{ q_{\phi}({z}|x,y)}~[\log {p_{\theta}({x}^{(i)}|{z})}+\log {p_{\theta}({y}^{(i)}|{z})}] 
	+ D_{KL}(q_{\phi}({z}|x,y)\parallel p_{\theta}({z})),\nonumber
\end{align}
Consequently, we can perform marginalization on $q_{\phi}({z}|{x},{y})$ to facilitate Bayesian inference for cross-view analysis.}

Obviously, this objective consists of two parts: (1). the reconstruction term, defined by expectation of  the data log-likelihood of the two views; (2). the prior regularization term, defined by the KL-divergence of the approximated posterior $q_{\phi}({z}|x,y)$ and the prior $p({z})$.

Based on Eq.~\eqref{eq:ICCA_F_q}, we propose Adversarial CCA, which achieves consistent encoding for multi-view data with the consistent constraint imposed on the marginalization of the implicit posteriors. 
\subsection{Design of ACCA} \label{sec:ACCA_design}
Graphical diagram of the ACCA is depicted in Fig.~\ref{fig:Graphics}.(d). Specifically, we adopt two schemes to achieve consistent encoding that promotes the multi-view data alignment.
\noindent \textbf{Holistic Encoding}: To facilitate the cross-view analysis task, we provide holistic information for the inference, i.e. $q({z}|{x},{y})$, $q({z}|{x})$ and $q({z}|{y})$, in ACCA.
\begin{remark}
	Based on Eq.\eqref{eq:ICCA_F_q}, we first explicitly model  $q({z}|{x},{y})$ by encoding an auxiliary view $XY$, which contains all the information of the two views. Then, we further adopt the two principle encodings $q({z}|{x})$ and $q({z}|{y})$ to facilitate the cross-view generation task. As holistic information is provided for the encoding of $z$, this scheme contributes to superior expressiveness of the common latent space.
\end{remark}
\noindent \textbf{Adversarial learning}:\label{sec:ACCA_design_adversarial} We then adopt an adversarial learning scheme to achieve consistency for three encodings by matching the marginalization of the implicit posteriors.
\vspace{-4mm}
\begin{remark}
	Since the prior regularization term in Eq.~\eqref{eq:ICCA_F_q} is still intractable for implicit posterior distribution, we design an adversarial learning scheme to achieve multi-view consistency by enforcing the approximation of the three marginalized posterior to the same implicit prior $p({z})$. Within this scheme, each encoder defines a marginalized posterior over ${z}$ (See Eq.(1) in~\citep{makhzani2015adversarial}).
		\begin{equation}
			q_{{x}}({z})=\int q({z}|{x})p({x})\,d{x},~~~~~~~~ q_{{y}}({z})=\int q({z}|{y})\,p({y})\,d{y},\nonumber
		\end{equation}
		\vspace{-3mm}
		\begin{equation}\label{eq:ACCA_three_agg}
			q_{{x},{y}}({z})=\iint q({z}|{x},{y})\,p({x},{y})d{x}d{y},
		\end{equation}
	\noindent As the true posterior $p({z}|{x},{y})$ is simultaneously approximated with the three terms, we adopt an generative adversarial network (GAN) that adopts the three inference model as generators and one shared discriminator to enforce the approximation of the three marginalized posteriors.
	\begin{equation}\label{eq:ACCA_three_appro}
		q_{{x}}({z}) \approx 
		q_{{y}}({z}) \approx
		q_{{x},{y}}({z}) \approx p({z}). 
	\end{equation} 
	Since the three marginalized posteriors are driven to match the prior $p({z})$ simultaneously, the adversarial learning scheme provides a consistent constraint for the incorporated encodings in ACCA. \QEDA
\end{remark}
\begin{figure}[t]
	\centering
	\hspace{-5mm}\includegraphics[width=11cm]{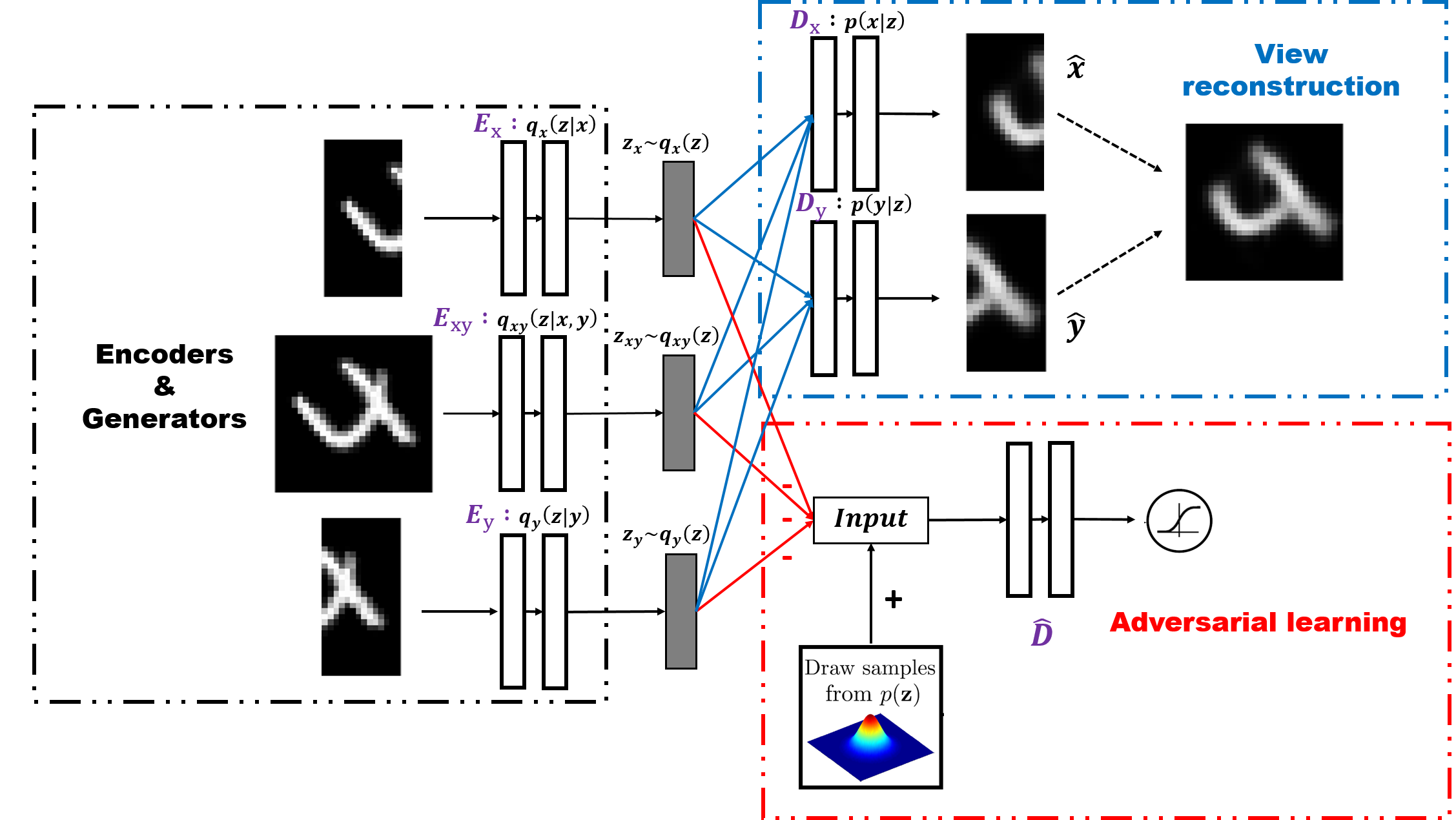}
	\caption{\label{fig:Network}  Overall structure of ACCA. \textbf{left panel}: The holistic encoding scheme. \textbf{Top of right panel}: The view reconstruction scheme; \textbf{Bottom of right panel}: The adversarial learning scheme.}
\end{figure}
\vspace{-4mm}
\subsection{Formulation of ACCA}\label{sec:ACCA_formulation}
Based on Eq.~\eqref{eq:ICCA_F_q} and the our design of ACCA, we formulate ACCA with the following equation and design the network structure as Fig.~\ref{fig:Network}.
\vspace{-1mm}
\begin{eqnarray}\label{eq:ICCA_ACCA}
	\lefteqn{\mathcal{L}_{\rm ACCA} (\Theta,\Phi; {x},{y}) = \frac{1}{N} \sum_{i = 1}^{N} [-
		\mathbb{E}_{ q_{\phi_{xy}}({z}|{x},{y})}[\log {p_{\theta_x}({x}|{z})}+\log {p_{\theta_y}({y}|{z})}]}\\
	&&\qquad \qquad \qquad \qquad \qquad \quad \:-\mathbb{E}_{ q_{\phi_{x}}({z}|{x})}[\log {p_{\theta_x}({x}|{z})}+\log {p_{\theta_y}({y}|{z})}]\nonumber\\ 
	&&\qquad \qquad \qquad \qquad \qquad\qquad \:  - \mathbb{E}_{q_{\phi_{y}}({z}|{y})}[\log {p_{\theta_x}({x}|{z})}+\log {p_{\theta_y}({y}|{z})}] + \mathcal{R}_{\rm{GAN}},\nonumber
\end{eqnarray}
where $\Theta$ and $\Phi$ in the left hand side denotes the parameters of the encoders and the decoders respectively, i.e. $\Theta$  = $\{\theta_{x},\theta_{y}\}$ and $\Phi$ = $\{\phi_{xy},\phi_{x},\phi_{y}\}$. 
\begin{remark}
	The framework of ACCA consists of 6 subnetworks (refer to Fig.~\ref{fig:Network}). The three encoders, $\{E_{x},E_{xy},E_{y}\}$, and the two decoders $\{D_{x},D_{y}\}$ compose the view-reconstruction scheme, which corresponds to the three terms in Eq.~\eqref{eq:ICCA_ACCA}. The three encoders also constitute a adversarial scheme with the shard discriminator $\hat{D}$. $\{E_{x},E_{xy},E_{y},\hat{D}\}$ provides a consistent constraint with $\mathcal{R}_{\rm GAN}$, namely
	\begin{align}
		\lefteqn{\mathcal{R}_{\rm GAN}(E_x,E_y,E_{xy},\hat{D})=\mathbb{E}_{{z^{'}}\sim p_({z})}\log(\hat{D}({z'}))+\mathbb{E}_{{z}_{{xy}}\sim q_{{x},{y}}({z}|{x,y})}\log(1-\hat{D}({z}_{{xy}}))}\nonumber
		\\
		&&\qquad\qquad\qquad\qquad\qquad+ \mathbb{E}_{{z}_{{x}}\sim q_{{x}}({z}|{x})}\log (1-\hat{D}({z}_{{x}}))+ \mathbb{E}_{{z}_{{y}}\sim q_{y}({z}|{y})}\log (1-\hat{D}({z}_{{y}}))\nonumber
	\end{align}
	Here we adopt $\{{z}_{{x}},{z}_{{y}}, {z}_{{xy}}\}$ to highlight ${z}$ obtained with different encodings. \QEDA
\end{remark}
\subsection{{Training}}\label{sec:ACCA_training}

For ACCA, we assume that each encoding, i.e. $q(z|*)$, to be a deterministic function of the corresponding view, and train the autoencoders and the adversarial networks jointly with ADAM Optimizer~\citep{radford2015unsupervised} in two phases - the reconstruction phase and the regularization phase - executed on each mini-batch. In the reconstruction
phase, the autoencoders update the encoders and the decoders to minimize the reconstruction error of the inputs in the three views. In the regularization phase, the adversarial network trained with the classic mechanism presented in~\citep{goodfellow2014generative}, with the generators (encoders) trained to maximize $\log D(z_{*})$ instead. Once the training procedure is done, the decoders of the autoencoder will define generative models
that map the imposed prior of $p(z)$ to the data distribution in each view.

\subsection{Connection with other CCA variants}\label{sec:ACCA_connections}

Facilitating Bayesian inference for implicit distributions based on marginalization in Eq.~\eqref{eq:ICCA_F_q}, our proposed ACCA can be connected with other existing CCA variants with certain assumptions on the posterior and likelihood distributions.

\vspace{0.5mm}
\noindent\textbf{\emph{Example 1: PCCA}}~\citep{bach2005probabilistic}.~~With an explicit conditional independent assumption for CCA, PCCA adopts Gaussian assumptions for both the likelihood and the prior, i.e. Eq.~\eqref{eq:PCCA}, to interpret linear CCA. Under the conditional independent constraint, the minimum CMI criterion, i.e.~$I({X};{Y}|{Z})=0$, is naturally satisfied. Then, due to the conjugacy of the prior and the likelihood, the posterior in Eq.\eqref{eq:ICCA_CMI_Fp} can be presented with an analytic solution, with which the model parameters can be directly estimated with EM algorithm. 

\vspace{0.5mm} 
\noindent\textbf{\emph{Example 2: MVAE}}~\citep{ngiam2011multimodal}. If we consider Gaussian models with \mbox{${z}\sim\mathcal{N}({\mu},{0})$}, $p_{\theta}(x|z) = N(g_{x}(z;\theta_{x}), I)$ and $ p_{\theta}(y|z) = N(g_{y}(z;\theta_{y}), I)$, we can see that the reconstruction term in Eq.~\eqref{eq:ICCA_F_q} measures the $l_{2}$ reconstruction error of the two inputs from the latent code $z$ through the DNNs defined with $g_{x}$ and $g_{y}$. {Without consistent constraint on the multi-view encodings, the objective of MVAE is given as   
\begin{eqnarray}
	\min \limits_{\theta,\phi} \frac{1}{2N} \sum_{i = 1}^{N} ~ {\| x^{(i)} - g_{x}(z;\theta_{x})\|}^{2} + {\| y^{(i)} - g_{y}(z;\theta_{y})\|}^{2}, \nonumber
\end{eqnarray}}
\noindent \textbf{\emph{Example 3: VCCA}}~\citep{wang2016deep}. Considering a model where the latent codes  ${z}\sim\mathcal{N}({\mu},\Sigma)$ and the observations $x|z$ and $y|z$ {both follow implicit distribution}, VCCA {adopts} variational inference to get the approximate posterior in Eq.\eqref{eq:ICCA_F_q} with two additional assumptions: \emph{1).} The input view at the test time can provide sufficient information for the multi-view encoding, {namely} $q_{\theta}(z^{(i)}|x^{(i)},y^{(i)})\approx {q_{\theta}(z^{(i)}|x^{(i)})}$; \emph{2).} The variational approximate posterior $q_{\theta}(z^{(i)}|x^{(i)}) \sim N(z^{(i)};\mu^{(i)},\Sigma^{(i)})$, where $\Sigma^{(i)}$ is a diagonal covariance, i.e. $\Sigma^{(i)} = \rm diag(\sigma_{1}^{2},\ldots ,\sigma_{d_{z}}^{2})$. In this case, {the} KL divergence term can be explicitly computed with $D_{KL}(q_{\phi}({z}^{(i)}|{x}^{(i)})\parallel p_{\theta}({z}^{(i)})) = -\frac{1}{2} \sum_{j = 1}^{d_{z}} (1+ \log \sigma_{ij}^{2}-\sigma_{ij}^{2}-\mu_{ij}^{2})$. As  $q_{\theta}(z^{(i)}|x^{(i)})$ is {confined with an explicit form}, Monte Carlo sampling~\citep{hastings1970monte} is adopted to approximate the expected log-likelihood term in Eq.\eqref{eq:ICCA_F_q}. {Drawing $L$ samples} $z_{i}^{l} \sim q_{\phi}({z}^{(i)}|{x}^{(i)})$, the objective of VCCA is given as
\begin{align}\label{eq:VCCA}
	&\min \frac{1}{N} \sum_{i = 1}^{N} [-\frac{1}{L} \sum_{l = 1}^{L}[\log {p_{\theta}({x}^{(i)}|{z}^{(i)}_{l})}+\log {p_{\theta}({y}^{(i)}|{z}^{(i)}_{l})}] 
	+ {D_{KL}(q_{\phi}({z}^{(i)}|{x}^{(i)})\parallel p_{\theta}({z}))}] \nonumber \\
	&\; s.t.\quad z^{(i)}_{l} = {\mu}^{(i)} + \Sigma^{(i)}\epsilon_{l},\ \text{where}\ \epsilon_{l} \sim\mathcal{N}({0},{I}_{d}),\ l= 1,\ldots,L. 
\end{align}
\textbf{\emph{Example 4: Bi-VCCA}}~\citep{wang2016deep}. As a variant of VCCA, Bi-VCCA adopts the encoding of both the two views, namely, $q_{\theta}({z}|{x})$ and $q_\theta({z}|{y})$ to solve the problem and its objective is given as a {heuristic} combination of Eq.~\eqref{eq:VCCA} derived with each encodings.
\begin{align}
	&\min \sum_{i = 1}^{N} [\frac{\lambda}{N} [ -\frac{1}{L} \sum_{l = 1}^{L}[\log {p_{\theta}({x}^{(i)}|{z}^{(i)}_{l_{x}})}+\log {p_{\theta}({y}^{(i)}|{z}^{(i)}_{l_{x}})}] 
	+{D_{KL}(q_{\phi}({z^{(i)}}|{x}^{(i)})\parallel p_{\theta}({z}))}] \nonumber \\
	& \quad + \frac{1-\lambda}{N} [-\frac{1}{L} \sum_{l = 1}^{L}[\log {p_{\theta}({x}^{(i)}|{z}^{(i)}_{l_{y}})}+\log {p_{\theta}({y}^{(i)}|{z}^{(i)}_{l_{y}})}] 
	+{D_{KL}(q_{\phi}({z^{(i)}}|{y}^{(i)})\parallel p_{\theta}({z}))}]] \nonumber \\
	&\; s.t.\quad z^{(i)}_{l_{x}} = {\mu_{x}}^{(i)} + \Sigma_{x}^{(i)}\epsilon_{l}, z^{(i)}_{l_{y}} = {\mu}_{y}^{(i)} + \Sigma_{y}^{(i)}\epsilon_{l},\ \text{where}\ \epsilon_{l} \sim\mathcal{N}({0},{I}_{d}),\ l= 1,\ldots,L, 
\end{align}
where $\lambda \in [0,1]$, which is the trade-off factor between the two encodings.

\section{Experiments}\label{sec:experiments}

In this section, we first verify that the minimum CMI criterion can be achieved with the optimization of ACCA. Next, we demonstrate the superiority of ACCA in handling implicit distributions with the prior specification. Then, we conduct correlation analysis to demonstrate the performance of ACCA in capturing nonlinear dependency. Then, we verify the {consistency for the inference of multi-view data} in ACCA through alignment analysis and cross-view generation.

\subsection{Verification of optimizing CMI}
We verify that ACCA achieves the proposed CMI criterion with one of the most commonly used multi-view learning dataset - MNIST left right halved dataset (MNIST\_LR)
\citep{andrew2013deep}. Details about the dataset and network design are in Table~\ref{tab:dataset}.

\begin{figure}[h]
	\centering
	\hspace{-2mm}
	\includegraphics[width=10cm]{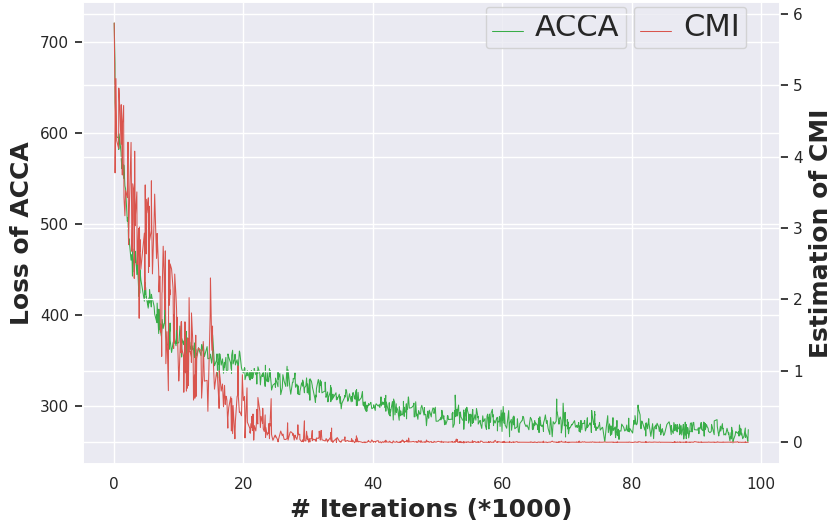}
	\caption{\label{fig:Convergence_2}  The CMI criterion is implicitly optimized with the objective of ACCA.} \label{fig:CMI_convergence}
	\vspace{-3mm}
\end{figure}

We estimate CMI the model training process with an open source non-parametric Entropy Estimation toolbox\footnote{https://github.com/gregversteeg/NPEET}. Fig.~\ref{fig:CMI_convergence} illustrates that the CMI gradually decreases during the training of ACCA and it reaches zero at a relatively early stage in the convergence of ACCA. The trend indicates two points:
\begin{itemize} 
	\item[1).] Instantiated with our objective for CCA based on CMI, {i.e. Eq.~\eqref{eq:ICCA_FN}}, ACCA implicitly minimizes CMI during the training process and the minimum CMI criterion can be achieved at the convergence of the objective.
	\vspace{1mm}
	\item[2). ] The conditional independent constraint (Eq.~\eqref{eq:conditional_independent}) is automatically satisfied, i.e. $I{({X};{Y}|{Z})=0}$, with the objective of ACCA.
\end{itemize}
\vspace{-2mm}
\subsection{Prior Specification}
We conduct correlation analysis on a toy dataset with specific prior to verify that ACCA benefits from handling implicit distributions in capturing nonlinear dependency.

%
\noindent \textbf{Toy dataset:} Analogizing to~\citep{DBLP:journals/nn/Hsieh00}, we construct a toy dataset that exists nonlinear dependency between the two views for the test. Specifically, let $X = W_{1}Z$ and $Y = W_{2}Z^{2}$, where $Z$ denotes a 10-D vector with each dimension $z\sim p(z)$, and $W_{1}\in \mathbb{R}^{10*50}$ , $W_{2}\in \mathbb{R}^{10*50}$ are the random projection matrices to construct the data. Details for the setting is presented in Table~\ref{tab:dataset}. As we consider nonlinear dependency with non-Gaussian prior, we set $p({z})$ with a mixture of Gaussian distribution in this experiment.
\begin{equation}
	z \sim p(z) = 0.2 * \mathcal{N}(0,\,1) + 0.5 * \mathcal{N}(8,\,2) + 0.3 * \mathcal{N}(3,\,1.5).\label{eq:zprior} \nonumber
\end{equation}

\noindent\textbf{Dependency metric:} Hilbert Schmidt Independence Criterion (HSIC)~\citep{DBLP:conf/alt/GrettonBSS05} is a state-of-the-art measurement for the overall dependency among variables. In this work, we adopt the normalized estimate of HSIC(nHSIC)~\citep{wu2018dependency} as the metric to measure the dependency captured by the embeddings of the test set ($Z_{X_{Te}}$ and $Z_{Y_{Te}}$) of each methods. The nHSIC computed with linear kernel and RBF kernel ($\sigma$ set with the F-H distance between the points) are both reported.
\newcommand{\tabincell}[2]{\begin{tabular}{@{}#1@{}}#2\end{tabular}} 

\begin{table}[t]
	\centering
	\large  
	\caption{Details of the datasets and network settings. 
	} \label{tab:dataset}
	\vspace{1mm}
	\setlength{\tabcolsep}{1.3mm}{
		\scalebox{0.64}{%
			\renewcommand{\arraystretch}{1.2}
			\begin{tabular}{|c|c|c|c|c|}
				\hline
				Dataset & Statistics & \tabincell{c}{\tabincell{c}{Dimension\\ of ${z}$}} & \tabincell{c}{Network setting (MLPs) \\ $\hat{D}= \{1024,1024,1024\}$} & Parameters \\ \hline
				\tabincell{c}{Toy dataset \\ (Simulated)} & \tabincell{c}{\# Tr= 8,000 \\ \# Te= 2,000} & d = 10 & \tabincell{c}{$E_{x}= \{1024,1024\}$; \\ $E_{xy}=\{1024,1024\}$; \\ $E_{y}=  \{1024,1024\}$}
				& \multirow{4}{*}{\tabincell{c}{ \\ For all the dataset: \\ learning rate = 0.001, \\  epoch = 100. \\ \\  For each dataset: \\ batch size tuned over \\ $\{32, 128, 256, 500, 512, 1000\}$;\\ $d$ tuned over $\{10, 30, 50, 70\}$}} \\ \cline{1-4}
				\tabincell{c}{MNIST L/R halved dataset \\ (MNIST\_LR) \\ \citep{andrew2013deep}} & \tabincell{c}{\# Tr= 60,000 \\ \# Te= 10,000} &  d = 30 & \tabincell{c}{$E_{x}= \{2308,1024,1024\}$; \\ $E_{xy}=\{3916,1024,1024\}$; \\ $E_{y}=  \{1608,1024,1024\}$} & \\ \cline{1-4}
				\tabincell{c}{MNIST noisy dataset \\ (MNIST\_Noisy) \\ \citep{wang2016deep}} & \tabincell{c}{\# Tr= 60,000 \\ \# Te= 10,000} & d = 50 & \tabincell{c}{$E_{x}= \{1024,1024,1024\}$; \\ $E_{xy}=\{1024,1024,1024\}$; \\ $E_{x}=  \{1024,1024,1024\}$} & \\ \cline{1-4}
				\tabincell{c}{Wisconsin X-ray \\ Microbeam Database \\ (XRMB) \\ \citep{wang2016deep}} & \tabincell{c}{\# Tr= 1.4M \\ \# Te= 0.1M} & d = 112 & \tabincell{c}{$E_x= \{1811,1811\}$; \\ $E_{xy}=\{1280,1280,1280\}$; \\ $E_y=  \{3091,3091\}$} & \\ \hline
	\end{tabular}}}
\end{table}
\vspace{1.5mm}
\noindent \textbf{Baselines:} We compare ACCA with several state-of-the-art generative CCA variants.
\begin{itemize}
	\item { \textbf{CCA}~\citep{hotelling1936relations}: Linear CCA model that learns linear projections of the two views that are maximally correlated.}
	\vspace{1mm}
	\item { 
		\textbf{PCCA}~\citep{bach2005probabilistic}: Probabilistic variant of linear CCA, which yields EM updates to get the final solution.  }
	\vspace{1mm}
	\item { 
		\textbf{MVAE}~\citep{ngiam2011multimodal}: Multi-View AutoEncoders, an CCA variant that discovers the dependency among the data via multi-view reconstruction.}
	\vspace{1mm}
	\item { 
		\textbf{Bi-VCCA}~\citep{wang2016deep}: Bi-deep Variational CCA, a representative generative nonlinear CCA model restricted with Gaussian prior.}
	\vspace{1mm}
	\item { \textbf{ACCA\_NoCV}: {An variant of ACCA which is designed without the encoding of the complementary view $XY$. This is compared to verify the efficiency of the holistic encoding scheme in ACCA.}}
	\vspace{0.5mm}
	\item { \textbf{ACCA(G)}; ACCA model implement with the standard Gaussian prior.}
	\vspace{1mm}
	\item { \textbf{ACCA(GM)}: ACCA model implement with a Gaussian mixture prior. The prior is set as the exact prior, {i.e. Eq~\eqref{eq:zprior}}, in the experiments.}\vspace{1mm}
\end{itemize}
As ACCA handles implicit distributions, which benefits its ability to reveal the intricacies of the latent space, higher nonlinear dependency is expected to achieve with this method. Table~\ref{hsic} presents the results of our correlation analysis. The table is revealing in several ways: 
\begin{itemize}
	\item[1).] Both CCA and PCCA achieve low nHSIC value on the toy dataset, due to their insufficiency in capturing nonlinear dependency.
	\vspace{1mm}
	\item[2).] The results of MVAE and Bi-VCCA are unsatisfactory. MVAE is influenced probably because it does not have the inference mechanism that qualifies the encodings. The Gaussian assumption made on the data distributions may also restrict its performance. Bi-VCCA suffers mainly because the heuristic combination of KL-divergence adopted for the approximation of the inferences causes inconsistent encoding problem. 
	\vspace{1mm}
	\item[3).] The three variants of ACCA all achieve good performance here. This indicates that the consistent constraint imposed on the marginalization in our model benefits the models' ability to capture nonlinear dependency.
	\vspace{1mm}
	\item[4).] ACCA (GM) archives the best result on both settings among our methods. This verifies that ACCA benefits from the ability to handle implicit distributions.
\end{itemize}

\subsection{Correlation Analysis}
We further test ACCA in capturing nonlinear dependency on three commonly used multi-view datasets, see (Table~\ref{tab:dataset}).
The result is presented in Table~\ref{hsic}. We can see that 
\begin{itemize} 
	\item [1).] Both of our models, ACCA(G) and ACCA (GM) achieve superb performance. The results of ACCA are much better than that of the state-of-the-art baselines. This indicates that the consistent constraint imposed on the marginalized posterior contributes to consistent encoding for the multi-view data which benefits for capturing dependency. 
	\vspace{1mm}
	\item [2).] ACCA(G) achieves better results than ACCA\_NoCV on most of the settings. This demonstrates that the adopted holistic encoding scheme also contributes to better dependency capturing ability for ACCA. 
\end{itemize} 
\begin{table}[t]
	\centering
	\renewcommand{\arraystretch}{1.2}
	\caption{The results of correlation analysis (best in bold).} \vspace{-2mm}
	\hspace{-2mm}
	\label{hsic}
	\scalebox{0.8}{
		\begin{tabular}{c|c|cccc|ccc}
			\toprule
			\textbf{Metric}&\textbf{Datasets} & \textbf{CCA} & \textbf{PCCA}& \textbf{MVAE} & \textbf{Bi-VCCA} &\textbf{ACCA\_NoCV} & \textbf{ACCA (G)} & \textbf{ACCA (GM)} \\ \hline
			\multirow{4}{*}{{\begin{tabular}[c]{@{}c@{}} \textbf{nHSIC}\\(linear kernel)\end{tabular}}}
			&toy        &0.0010   &0.1037&0.1428& 0.1035&0. 8563&  0.7296& 
			\textbf{0.9595}\\ 
			&MNIST\_LR  & 0.4210  &0.3777&0.2500&0.4612 & 0.5233 & 0.5423 & \textbf{0.6823}\\
			&MNIST\_Noisy&0.0817 &0.1037&0.4089 &0.1912 &0.3343 & 0.3285& \textbf{0.4133}\\
			&XRMB       &  0.1735 &0.2031&0.3465&  0.2049& 0.2537 & 0.2703& \textbf{0.3482}\\ \hline
			\multirow{4}{*}{{\begin{tabular}[c]{@{}c@{}} \textbf{nHSIC}\\(RBF kernel)\end{tabular}}}
			&toy         &0.0029 &0.2037& 0.2358 & 0.2543 & 0.8737 & 0.5870 & 
			\textbf{0.8764}\\ 
			&MNIST\_LR   & 0.4416   &0.3568& 0.1499  &0.3804& 0.5799 & 0.6318 & \textbf{0.7387}\\
			&MNIST\_Noisy & 0.0948  &0.0993& 0.4133&0.2076 &0.2697 & 0.3099& \textbf{0.4326}\\
			&XRMB       &  0.0022  &0.0025& 0.0022 & 0.0027  & 0.0031 & 0.0044& \textbf{0.0058}\\
			\bottomrule
	\end{tabular}}
\end{table}

\subsection{Verification of consistent encoding}
In this subsection, we specially verify the consistent encoding of ACCA with MNIST\_LR dataset.  We first illustrate the approximation of the three encodings in ACCA. Then, we conduct alignment verification to verify the effect of consistent encoding. 

\subsubsection{\mbox{Approximation of three encodings}}

\begin{figure}[h]
	\centering
	\includegraphics[width=10cm]{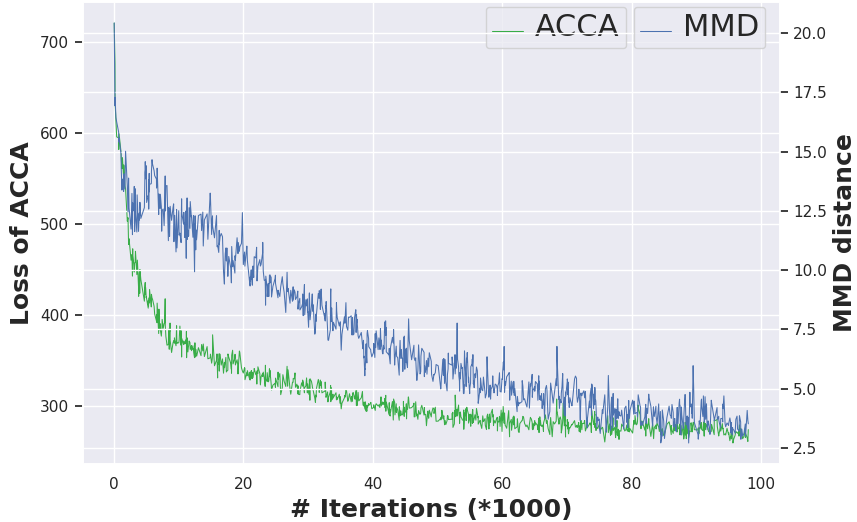}
	\vspace{2mm}
	\caption{\label{fig:MMD}  Verification for consistent encoding of ACCA, i.e. Eq.~\eqref{eq:ACCA_three_appro}.}
\end{figure}

We adopt MMD distance to verify the approximation of the three encodings in ACCA. Specially, we calculate the sum of the MMD distance between the three encodings and the prior $p(z)$ in Eq.~\eqref{eq:ACCA_three_agg} during the training process. Fig.~\ref{fig:MMD} shows that result decreases during the convergence of ACCA. This verifies that ACCA achieves Eq.~\eqref{eq:ACCA_three_appro}, i.e. the consistent encoding during the optimization.
\subsubsection{Alignment verification}
We first embed the data into two-dimensional space to verify the alignment of the multi-view data in the common latent space. 

\begin{figure}[t]
	\centering
	\includegraphics[width=12cm]{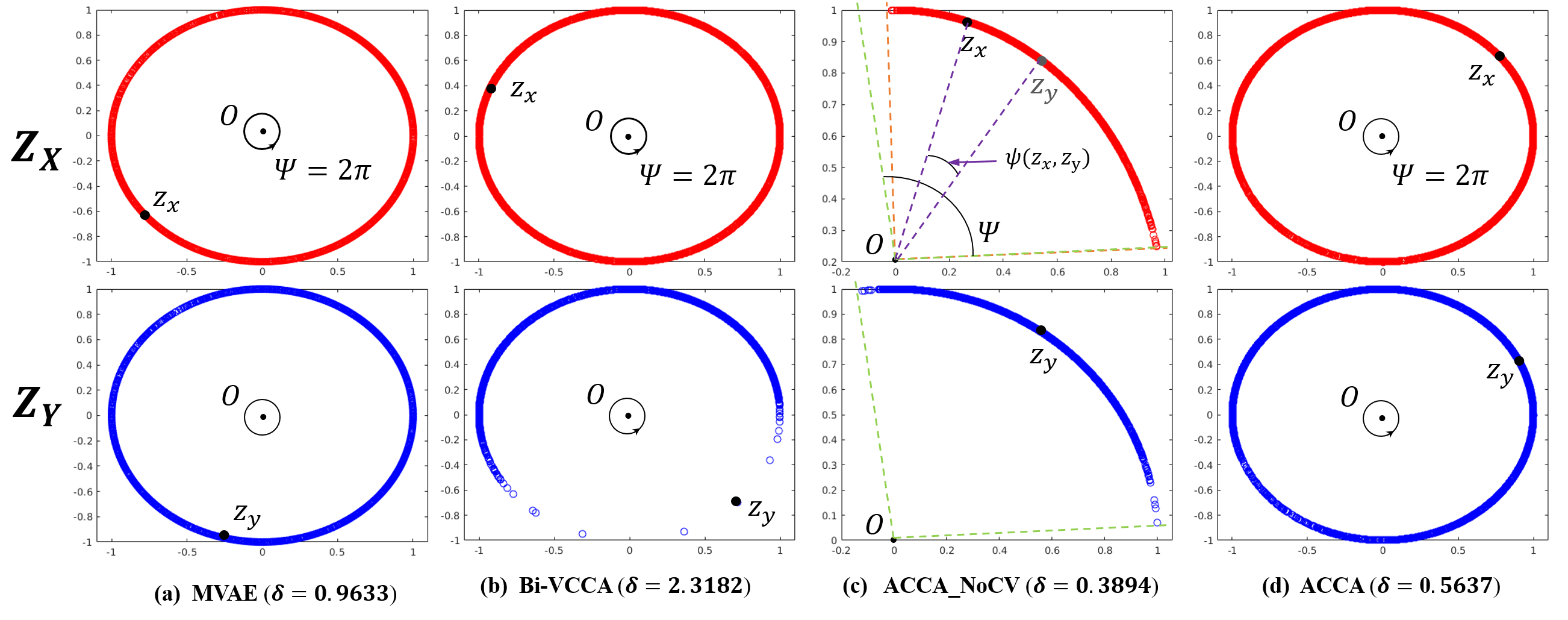}
	\caption{\label{fig:Align}  Visualization of the embeddings obtained for the two views, with each row represents the embeddings obtained with view $X$ and view $Y$, respectively. (${z}_{x}$, ${z}_{y}$) denote a pair of correspondent embedding. $\delta$ indicates the misalignment degree of each method. Methods with a smaller value of $\delta$ are better.}
\end{figure}

Specifically, projecting the paired testing data to the latent space with Gaussian prior, we take the origin $O$ as the reference and adopt angular difference to measure the distance of the paired embeddings, i.e. $\phi ({z}_{x},{z}_{y}) = \angle {z}_{x}O{z}_{y}$ (see Fig.~\ref{fig:Align}.(b)). The misalignment degree of the multi-view is given by
\begin{equation}
	\delta = \frac{\sum_{1}^{N} \psi({z}_{x},{z}_{y})} {N*\Psi},   
\end{equation}
Where $\Psi$ denotes the maximum angle of all the embeddings and $N$ is the number of data pairs. 

We compare ACCA with MVAE, Bi-VCCA and ACCA\_NoCV here, as they have encoding for both the two views. The results are presented in Fig.~\ref{fig:Align}. We have the following observations.
\begin{itemize} 
	\item[1).] The regions for the paired embeddings of Bi-VCCA are even not overlapped and the misalignment degree of Bi-VCCA is $\delta = 2.3182$, which is much higher than the others. This shows that Bi-VCCA greatly suffers from the misaligned encoding problem for the multi-views due to the {inferior inference of the multi-view data}. 
	\vspace{1mm}
	\item[2). ] Our two methods, ACCA and ACCA\_NoCV, achieves superior alignment performance compared with MVAE (small misalignment degree) and Bi-VCCA (better-overlapped regions and small misalignment degree). This shows the effectiveness of the consistent constraint on the marginalization for view alignment in ACCA. 
	\vspace{1mm}
	\item[3). ] The embeddings of ACCA are uniformly distributed in the latent space compared with that of ACCA\_NoCV, indicating that the complementary view, $XY$ provide additional information for the holistic encoding, which benefits the effectiveness of the common latent space. 
\end{itemize} 
\subsection{Cross-view generation}
We apply ACCA to novel cross-view image generation task, where consistent encoding is critical for the performance. Specifically, this task aims at whole-image recovery given the partial input image as one of the views. We conduct the experiment on MNIST and CelebA~\citep{liu2015deep}, which are both commonly used for image generation. We add noise to the original MNIST to verify the robustness of ACCA. {Specifically,} we divide the test data in each view into four quadrants and masked one, two or three quadrants of the input with grey color~\citep{sohn2015learning} {and use the noisy images as the input for test.} The result is evaluated from both qualitative and quantitative aspects. For CelebA, we halved the RGB images into top-half and bottom-half and designed CNN architecture for this problem. Details for the network design is given in Table.~\ref{tab:dataset_celebA}.
The evaluation is conducted on the quality of the generated image, e.g. is the image blurred, does the image shows clear misalignment at the junctions in the middle. {Since MVAE and Bi-VCCA are the baseline methods that can support this cross-view generation task,} we compare the two methods here. 

\subsubsection{MNIST}

\begin{figure}[t]
	\centering
	\hspace{-5mm}
	\includegraphics[width=11cm,height= 8.5cm]{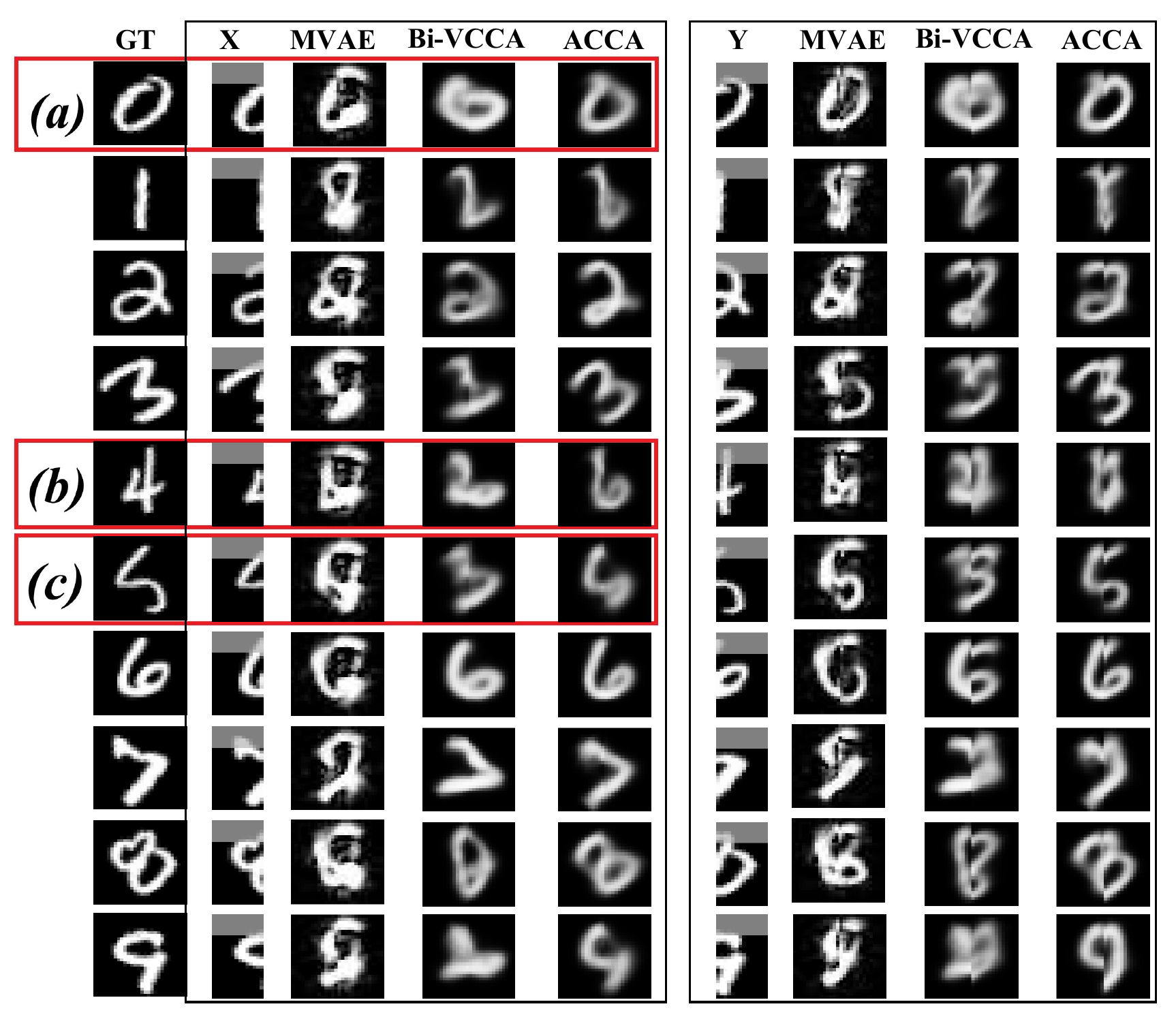}
	\caption{ Generated samples given one quadrant noisy image as input. The first column is the ground truth. The next three columns show the input for view $X$ and the generated image with Bi-VCCA and ACCA, respectively. The last three columns are that of $Y$.}\label{fig:qualitative-analysis}
\end{figure}

\textbf{Qualitative analysis}: Fig.~\ref{fig:qualitative-analysis} presents some of the recovered images (column 3-5) obtained with 1 quadrant input. This figure clearly illustrates that, given the noisy input, the images generated with ACCA is more real and recognizable than that of MVAE and Bi-VCCA.
\vspace{-3mm}
\begin{table}[h]
	\centering
	\renewcommand{\arraystretch}{1.2}
	\caption{ Pixel-level accuracy for full image recovery with masked inputs for different input views on the MNIST dataset.}
	\label{pixel-level-accuracy}
	\scalebox{1}{
		\begin{tabular}{c|c|ccc}
			\hline
			\multirow{2}{*}{\begin{tabular}[c]{@{}c@{}}\textbf{Input}\\ (halved image)\end{tabular}} & \multirow{2}{*}{\textbf{Methods}}& \multicolumn{3}{c}{\textbf{Gray color overlaid}} \\ \cline{3-5} 
			& & 1 quadrant & 2 quadrants & 3 quadrants \\ \hline
			\multirow{3}{*}{\textbf{Left}} & MVAE & 64.94 & 61.81 & 56.15\\ 
			& Bi-VCCA & 73.14 & 69.29 & 63.05 \\ 
			& ACCA & \textbf{77.66} & \textbf{72.91} & \textbf{67.08} \\ \hline
			\multirow{3}{*}{\textbf{Right}} & MVAE & 73.57 & 67.57 & 59.69 \\ 
			& Bi-VCCA & 75.66 & 69.72 & 65.52 \\
			& ACCA & \textbf{80.16} & \textbf{74.60} & \textbf{66.80} \\ 
			\bottomrule
	\end{tabular}}
\end{table}
\vspace{-3mm}
\begin{itemize}
	\item [1).] The image generated with MVAE shows the worst quality. The images contain much noise compared with other methods. In many cases, the "digit" is hard to identify, e.g. case (b). In addition, the generated image of MVAE shows clear misalignment at the {junctions of the halved images}, e.g. case (a).
	\vspace{1mm}
	\item [1).] The images generated by Bi-VCCA are much more blurred and less recognizable than that of ACCA, especially in case (a) and case (b).
	\vspace{1mm}
	\item[2).] ACCA can successfully recover the noisy half images which are even confusing for our human to recognize. For example, in case (b), the left-half image of digit ``5'' looks similar to the digit ``4'', ACCA succeeds in recovering the true digit.
\end{itemize}

\textbf{Quantitative evidence}: We compare the pixel-level accuracy of the generated images in Table~\ref{pixel-level-accuracy}. It shows that our ACCA consistently outperforms Bi-VCCA given the different level of masked input images. It is also interesting that using the left-half images as the input tends to generate better images than using the right-half. It might be because of the right-half images contain more information than the left-half images, which will result in better network training for more accurate image generation.
\subsubsection{CelebA}
\begin{figure}[t]
	\centering
	\hspace{-3mm}
	\includegraphics[width=12cm,height= 14cm]{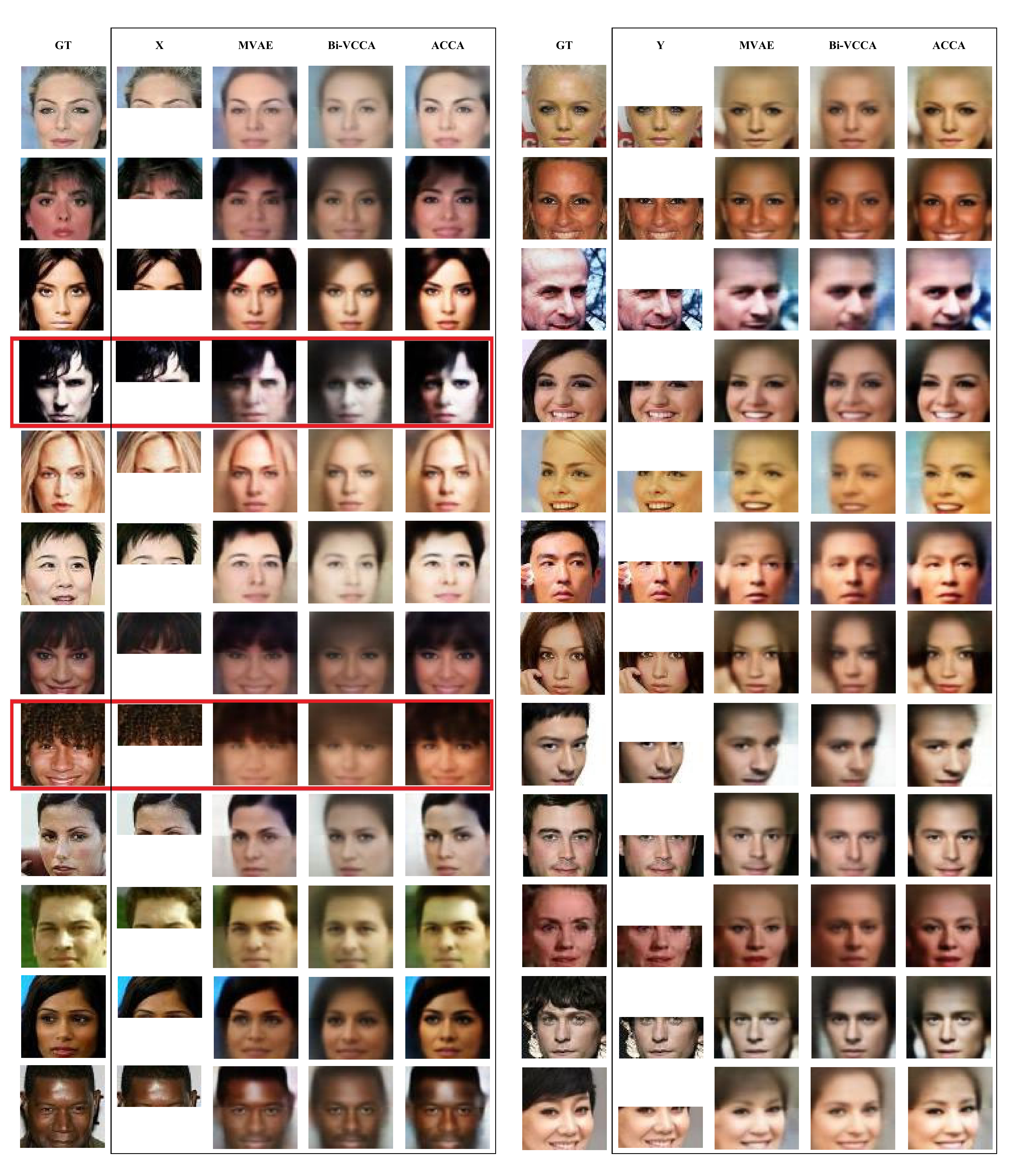}
	\caption{ The image generated with different methods on CelebA. }\label{fig:CelebA}
	\vspace{-4mm}
	\end{figure}
Fig.~\ref{fig:CelebA} shows the image samples generated for CelebA dataset. We have mainly two observations.
\begin{itemize}
	\item [1).] The samples generated with MVAE shows clear misalignment at the junctions, especially when the images are with colored backgrounds. Some of the images are much blurred to see the details, e.g. the samples circled with red.
	\vspace{1mm}
	\item [2).] The samples generated by Bi-VCCA are commonly blurred than the other two. The observation is quite obvious in the image generated with the top-half image, which contains much less details than the bottom-half image.
	\vspace{1mm}
	\item[3).] The images generated with ACCA show better quality compared with the others, considering both the clarity and the alignment of junctions.
\end{itemize}
\begin{table}[t]
	\centering
	\large  
	\caption{Details of the cross-view generation on CelebA dataset.} \label{tab:dataset_celebA}
	\vspace{1mm}
	\setlength{\tabcolsep}{1.3mm}{
		\scalebox{0.62}{%
			\renewcommand{\arraystretch}{1.2}
			\begin{tabular}{|c|c|c|c|c|c|c|}
				\hline
				Dataset & Statistics & \tabincell{c}{\tabincell{c}{Dimension\\ of ${z}$}} & \multicolumn{2}{c|}{\tabincell{c}{Architecture
						\\(Conv all with batch normalization before LReLU)}} &Parameters\\ \hline
				\tabincell{c}{CelebA\\\citep{liu2015deep} } & \tabincell{c}{\# Tr= 201,599 \\ \# Te= 1,000} & d = 100 &\multicolumn{1}{c|}{\tabincell{c}{\textbf{Encoders}:\\Conv 128*5*5 (stride 2),\\Conv 256*5*5 (stride 2),\\Conv 512*5*5 (stride 2);\\dense: 100.\\ \textbf{Decoders}:\\dense: 8192, relu;\\dConv: 512*5*5 (stride 2),\\dConv: 256*5*5 (stride 2),\\dConv: 128*5*5 (stride 2),\\dConv: 3*2*5,  (stride1*2);\\ Tanh.
				}}&\tabincell{c}{\textbf{Discriminator: $\hat{D}$}:\\dense: 128$\rightarrow$64$\rightarrow$1,\\
					sigmoid.\\}& \tabincell{c}{Epoch = 10;\\ Batchsize = 64;\\ lr= 0.0002;\\ Beta1 = 0.05;
				}\\ \hline
	\end{tabular}}}
\end{table}
\section{Conclusion} \label{sec:conclusions}
In this paper, we present a probabilistic interpretation for CCA based on implicit distributions. Our study discusses the restrictive assumptions of existing CCA variants and provides a unified probabilistic interpretation for these models based on Conditional Mutual Information. We present minimum CMI as a new criterion for CCA and provides an objective which can provide efficient estimation for CMI without explicit distributions. We further propose Adversarial CCA which shows superior performance in both nonlinear correlation analysis and cross-view generation, due to the consistent constraint imposed on marginalizations. Due to the matching of multiple encodings, ACCA can also be adopted to other practical tasks, such as image captioning and translation. Furthermore, because of the flexible architecture designed in~Eq.\eqref{eq:ICCA_ACCA}, proposed ACCA can be easily extended to multi-view task of $n$ views, with ($n+1$) encoders and ($n$) decoders.

\bibliographystyle{spbasic}      
\bibliography{ICCA}   

\end{document}